# Deep Learning Architect: Classification for Architectural Design through the Eye of Artificial Intelligence

Yuji Yoshimura, Bill Cai, Zhoutong Wang, and Carlo Ratti

## Abstract

This paper applies state-of-the-art techniques in deep learning and computer vision to measure visual similarities between architectural designs by different architects. Using a dataset consisting of web scraped images and an original collection of images of architectural works, we first train a deep convolutional neural network (DCNN) model capable of achieving 73% accuracy in classifying works belonging to 34 different architects. Through examining the weights in the trained DCNN model, we are able to quantitatively measure the visual similarities between architects that are implicitly learned by our model. Using this measure, we cluster architects that are identified to be similar and compare our findings to conventional classification made by architectural historians and theorists. Our clustering of architectural designs remarkably corroborates conventional views in architectural history, and the learned architectural features also coheres with the traditional understanding of architectural designs.

___
Y. Yoshimura (Corresponding author) • B. Cai • C. Ratti
SENSEable City Laboratory, Massachusetts Institute of Technology, 77 Massachusetts Avenue, Cambridge, MA 02139, USA
Email: yyoshi@mit.edu

Z. Wang
Department of Architecture, Harvard GSD, 48 Quincy, St. Cambridge,
MA 02138, USA

# 1. Introduction

This paper proposes to classify architectural designs through computer vision techniques, purely based on their visual appearances. The question asked is whether or not state-of-the-art deep learning techniques can identify distinguishing design features of each architect and cluster them in a similar way to the one of architectural historians and theorists. Our hypothesis is that the internally learnt discriminative factors provides us the key to explaining the difference in designs between architects rather than merely recognizing the architectural elements. Thus, we try to isolate the visual factor from others (i.e., prior knowledge, memory, image) for the purpose of the classification of architectural designs.

Architectural history and theory classify architectural styles and types from various perspectives (see Forty, 2000, pp. 304-311). A style of architecture (i.e., Renaissance, Baroque) provides a basic format for designing an individual building in a geographical region during a specific epoch. The ornaments that pertain to the specific style are considered as the expression of beauty for each age; these features then convert ordinary buildings into structures of architectural significance. Thus, visual elements such as windows, pillars, or architectural orders (Onians, 1988) can provide a clue for identifying and classifying its architecture into a specific style.

Conversely, compared with the element-based classification for the historical types of architecture, most classifications for modern and contemporary architecture are largely either function based or building-type based (Rossi, 1960, p.36). This is largely due to changes in design concept. The international style (Hitchcock and Johnson, 1932) aims to express the function of the building through a "machine aesthetic", resulting in the shaping of modern architecture into a white cube. The historical ornament and decoration are rejected and "the machine" became the model for the modern architecture. Thus, the modernists tend to reduce any forms to abstraction (Frampton, 1992, p.210). In addition, space and its experience become one of the most important topics in the design of modern and contemporary architecture. This makes further complicates the classification because space cannot be identified by elements; rather, it appears when it is enclosed by the combination of several spatial elements together with light. Consequently, the classification comes to rely more on an abstract and dematerialized concept rather than being based on elements, as it did in previous periods.

In order to fill in this gap, this paper proposes a computational approach to classify designs of modern and contemporary architecture. The objective of this paper is twofold: (1) we present our analytical framework as the research methodology and (2) we show our preliminary result for our current research. Our proposed methodology is based on the machine eye rather than the human eye, which provides us with a different perspective and insights. From a technical point of view, our contribution includes the application of DCNN to the design style classification, because it is a different topic from the element-based classification. The former is well researched, including the field of architecture (Shalunts et al., 2011), while most of the latter is researched in the fields of art (Tan et al., 2016). For this purpose, we employ recently developed deep learning techniques in processing the visual images to classify the given datasets through the training samples. The obtained results are clustered depending on the visual similarities measured by the algorithm. The final results are compared with the classifications made by architectural historians and theorists. Thus, we show the capability of artificial intelligence to classify designs in modern and contemporary architecture. The machine-eye-based classification provides us with some insights to enhance our understanding of architectural design.

The paper is structured as follows: Section 2 provides a literature review and describes the analytical methodology for this paper. Section 3 describes the dataset used in our study. Section 4 presents our study and the preliminary results. We conclude in section 5, suggesting future work.

## 2. Related works and analytical methodology

### 2.1 Related works

The analytical methodology of this paper relies on deep convolutional neural networks (DCNNs), which have recently achieved a remarkable performance in the fields of image classification (Krizhevsky et al., 2012), scene recognition (Zhou et al., 2014), speech recognition (Abdel-Hamid et al., 2012), and machine translation (Bahdanau et al., 2015). The main advantage of DCNN methods over traditional computer vision and machine learning techniques is their ability to identify and generalize important features, and employ these learned features to classify objects according to their appropriate labels. The visual features are engineered and extracted, creating high-level and semantic features of the input images without human intervention.

**Table 1.** Summary of previous literature on classification of visual elements in art, architecture, and urban studies.

| | Objective | Model | Dataset |
|---|---|---|---|
| Elgammal et al. (2018) | Characteristics of style in art and patterns of style changes | AlexNet, VGGnet, ResNet, and variants | 76,921 paintings from 1,119 artists with 20 classes from WikiArt. 1,485 images of paintings from Artchive dataset with 60 artists for visualization and analysis. |
| Obeso et al. (2017) | Classification of Mexican heritage buildings' architectural styles | GoogLeNet and AlexNet for a Saliency-Based and a Center-Based approach | 16,000 labeled images in 4 categories, out of which 3 are Mexican buildings (pre-Hispanic, colonial, modern) and one "other." |
| Llamas et al. (2017) | Classification of architectural heritage elements | AlexNet and Inception V3 for CNN, ResNet and Inception-resNet-v2 for Residual Networks | More than 10,000 images classified into 10 types of architectural elements, mostly churches and religious temples. |
| Zhang et al. (2018) | Prediction of urban elements that cause human perceptions | DCNN, PSPNet | 1,169,078 images from MIT Place Pulse for training a DCNN model. 245,388 images from Google Street View from Shanghai and 135,175 from Beijing to predict human perception. |
| Cai et al. (2018) | Quantification of street-level urban greenery | PSPNet (Pyramid Scene Parsing Network) and ResNet for DCNN semantic segmentation | 500 street images from Google Street View and 500 images of cityscapes from vehicle-mounted cameras. |

Table 1 presents the summary of previous literatures on classification of images in art, architecture, and urban studies. In the art field, many studies deal with the classification of artistic styles and artists using low-level features-based approaches (Li et al., 2012; Saleh et al., 2016; Tan et al., 2016; Elgammal et al., 2018). Li et al. (2012) propose an edge detection and clustering-based segmentation to extract the characteristics of van Gogh's brushstrokes and distinguish the artist from others. Saleh et al. (2016) attempt to identify similarities in artists' works and explore the influence and connections between artists. Elgammal et al. (2018) explore factors which make art style change in learnt discriminative features by the machine.

Conversely, architecture studies propose to classify historical architecture into styles based on historical architectural elements (Shalunts et al., 2011, 2012; Goel et al., 2012; Zhang et al., 2014; Lee et al., 2015; Shalunts, 2015; Llamas et al., 2017; Obeso et al., 2017). Others focus on ordinary buildings dispersed in the city to identify the urban elements which are the determinant factors of each city (Doersch et al., 2012; Lee et al., 2015). Doersch et al. (2012) explore the urban elements that appear frequently in a geographically determined location but do not appear in other areas, while Lee et al. (2015) attempts to identify the visual features which specify the architectural styles for each period and the evolution of architectural elements over time.

Although the classifications of historical architecture, including buildings, monuments, and cultural heritage, are well researched, there have been few attempts to classify modern and contemporary architecture or architects. This shortage may derive from the difficulty in identifying the features of space, which cannot be identified by elements. Space is not material, rather it appears when it is enclosed by a combination of several spatial elements. In addition, space is experienced not only through our perceptions but also through our other senses. In terms of techniques, most previous literature employs clustering and learning of local features (Shalunts et al., 2011), but not deep learning (Llamas et al., 2017).

This paper attempts to classify designs of modern and contemporary architecture using a deep convolutional neural network. We try to capture spatial design features rather than recognize specific visual features of buildings (i.e., window, domes, pillars). Our approach is similar to the artistic style classification, in which recognizing an artistic style is a different topic from identifying elements (Saleh et al., 2016; Elgammal et al., 2018), because style is independent from the content of a drawing. Thus, we explore the learnt internal discriminative factors to explain modern and contemporary architecture and its space.

### *2.2 Deep convolutional neural network*

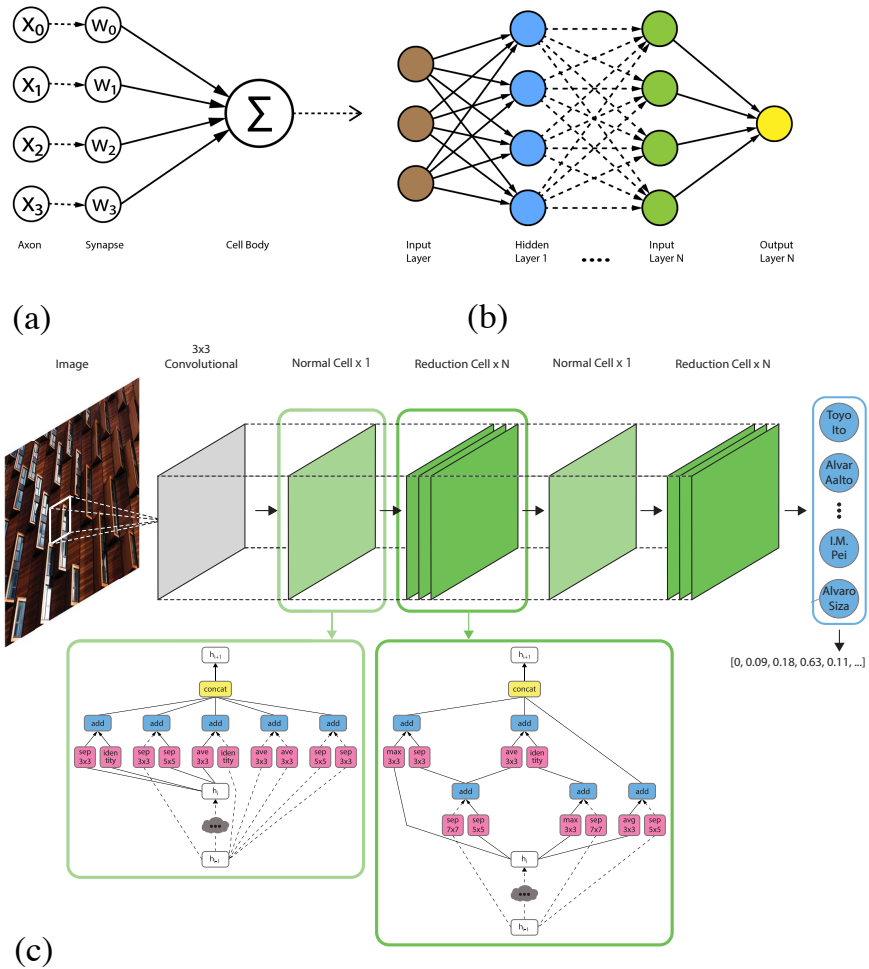

**Figure 1.** (a) Diagram of the neuron. (b) Diagram of a fully connected neural network with N layers. The input layer (zero layer) has three neurons and the hidden layers has four neurons. (c) Two repeated motifs termed "normal cell" and "reduction cell", discovered as the best convolutional cells in the CIFAR-10 dataset, modified the Figure 4 in Zoph and Shlens, (2018).

The deep convolutional neural network (DCNN) is a class of deep, feedforward artificial neural networks mainly applied in the analysis of visual imagery. Figure 1 (a) shows a simple diagram of the reproduction of the neuron's system in DCNN. Input data $x$ is multiplied by weight $w$, to which bias *b* is also added, for function *f*, producing *y* as the output. Several neurons are combined to create the neural network (see Figure 1 (b) for the diagram of the neural network). In order to be effective, a neural network has to discover the optima weights for all the connections in the network. Similar to the human brain, which changes the strength of connection between synapses, the neural network adjusts the weights through the learning process and seeks the best combination of weights that minimize the error between the correct classification of an input and the output of the network at the last layer.

DCNNs stack many convolutional layers into a single network. Convolutional layers allow for dimensional reduction in high-dimensional problems and have driven recent success in object detection, classification, and segmentation (Krizhevsky et al., 2012). Multiple stacked convolutional layers allow DCNNs to learn feature hierarchies, beginning from simple edges and shapes in the early layers, and ending with complex semantic features such as windows and roofs (Girshick et al., 2014).

In our experiment, we utilized NASNet, a novel program that achieves state-of-the-art accuracy while halving the computational cost of the best reported results (Zoph and Shlens, 2018). NASNet is composed of two types of layers: normal layer and reduction layer (Figure 1 (c)), both designed by auto machine learning.

### 2.3 Visual explanation of DCNN

In a stacked convolutional neural network model, each layer contains increasingly complex features and is optimized to identify distinguishing traits of architects. Consequently, the numerical matrix representing the weights in the last activation layer of the DCNN models represents high-level visual concepts that help to distinguish between architects.

We employed gradient-weighted Class Activation Mapping (Grad-CAM) (Selvaraju et al., 2017) to examine the visual explanation of NASNet. It clarifies the influential gradients and their regions with respect to the output of NASNet. Compared with another popular visualization technique such as Class Activation Mapping (Zhou et al., 2014), Grad-CAM combines feature maps using a gradient signal that does not require any modification in the network architecture, thus making it possible to be applied to NASNet. To compute Grad-CAM, we used the following formulas proposed by Selvaraju et al. (2017):

$$L^c_{Grad-CAM} \in R^{u \times v} \qquad (1)$$

$$\alpha^c_k = \frac{1}{Z}\sum_i \sum_j \frac{\partial y^c}{\partial A^k_{ij}} \qquad (2)$$

$$L^c_{Grad-CAM} = ReLU(\sum_k \alpha^c_k A^k) \qquad (3)$$

The objective function in this task is defined as (1), where the width *u* and height *v* for any class are. First, we compute $\alpha^c_k$, the global average pooling by (2), in which $A^k_{ij}$ indicates the element in matrix *ij* of kth feature map, $A^k_{ij}$ is the output of the feature map A, and Z is the normalized item. Second, we compute $L^c_{Grad-CAM}$, the heat map of Grad-CAM by summing up the feature maps $A^k$ weighted by $\alpha^c_k$. *ReLU* computes the pixel to increase the output of $Y_C$ (see Selvaraju et al., 2017, for the technical description in more detail).

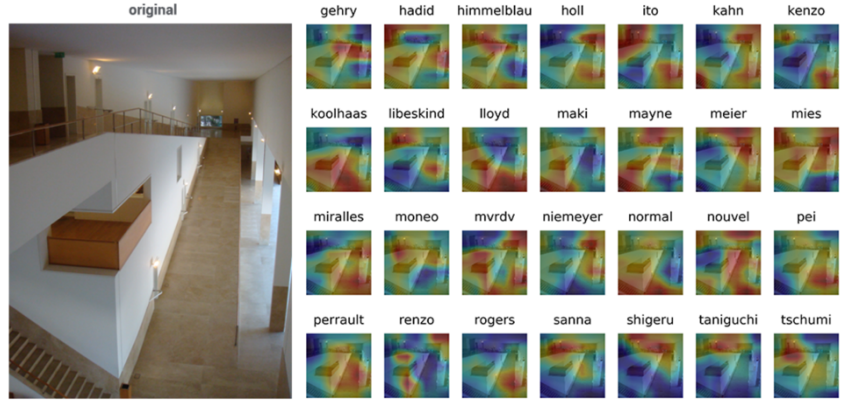

**Figure 3.** Grad-CAM applied to an image taken from Alvaro Siza's work

Figure 3 presents an example of Grad-CAM, applied to an image taken from Alvaro Siza's work. It enables us to understand the focus of the machine eye for the classification of objects.

### *2.4 Dimension reduction and clustering*

Using the outputs of the last layer, which is the product of weights and the outputs of the second-to-last layer (deep feature), we are able to cluster and measure the similarities between the visual signatures used to distinguish different architects by using linear principal component analysis (PCA) and kernel PCA (Jolliffe, 2002). To carry out the PCA, we take and normalize the last layer before softmax for *n* images in DCNN. Then we can construct $D = [d_1, d_2, \ldots, d_n]$ matrix $D \in R^{d \times n}$, where $d$ denotes the number of categories, and $n$ denotes the number of images. Let the first $k$ principal components of $D$ be $B = [b_1, b_2, \ldots, b_k]$. In our case, k = 2. The objective function is:

$$\max_{b} \sum_{i=1}^{n} \| B^T d_i \|_2^2 = B^T \Gamma B \qquad (4)$$

with constraint $B^T B = I$. We use least-squares estimation to minimize the objective function:

$$E_{pca}(B) = \sum_{i=1}^{n} e_{pca}(e_i) = \sum_{i=1}^{n} \| d_i - BB^T d_i \|_2^2 \qquad (5)$$

After dimension deduction, we use k-means to find clusters among different architects.

## 3. Data collecting and sampling

We used a combination of private collection of photographs and public collection of images found via the internet. We chose works of 34 architects, most of whom are the past Pritzker Prize recipients, which are considered to have specific and distinguished architectural design features. We also included categories of typical residential houses to allow the trained model to better differentiate between general buildings and works from renowned architects. Moreover, this sample class enabled us to measure how well known architectural designs are from typical designs.

**Table 2.** Architect and sample size of collected photographs

| Architect | Sample Size | Architect | Sample Size |
|---|---|---|---|
| Alvar Aalto | 460 | Oscar Niemeyer | 437 |
| Alvaro Siza | 1,289 | Peter Eisenman | 331 |
| Bernard Tschumi | 288 | Rafael Moneo | 278 |
| Coop Himmelblau | 390 | Rem Koolhaas | 373 |
| Le Corbusier | 527 | Renzo Piano | 542 |
| Daniel Libeskind | 406 | Richard Meier | 464 |
| Dominique Perrault | 234 | Richard Rogers | 406 |
| E. Souto de Moura | 559 | SANNA | 393 |
| Enric Miralles | 518 | Shigeru Ban | 216 |
| Frank Gehry | 669 | Steven Holl | 498 |
| Frank Lloyd Wright | 1,177 | Tadao Ando | 730 |
| Fumihiko Maki | 457 | Kenzo Tange | 454 |
| I.M. Pei | 419 | Thom Mayne | 723 |
| Jean Nouvel | 358 | Toyo Ito | 672 |
| Louis Kahn | 1,442 | Yoshio Taniguchi | 528 |
| Mies van der Rohe | 881 | Zaha Hadid | 635 |
| MVRDV | 253 | Normal house | 1,305 |
| Norman Foster | 256 | Total | 19,568 |

To collect images from the internet, we created several combinations of keywords relevant to a specific architect (i.e., name of the architect, type of architecture, etc.). After collecting the raw dataset, we manually cleaned it by eliminating the mislabeled and unclear images from the samples. We also added photographs of specific works by some of the architects which were personally taken by the authors of this paper. As a result, the total number of the collected samples is 19,568 (see Table 2 for architects and the corresponding sample size). All images were annotated with the id of the architect who was responsible for the design.

We implemented our DCNN model using the Google TensorFlow library and the algorithms implemented by Python. The computer had a Linux system (Ubuntu 16.04) with an Intel Core i5, CPU, 16GB memory and two parallel GeForce GTX 1070Ti. The training was completed in 8 hours.

## 4. Results

This section presents the results of our proposed methodology. First, we examine the overall model accuracy and comparisons between different architects and between different types of images. Second, we present the Grad-CAM generated heat maps, which were used to analyze the point where the model focuses on the picture during the identification process. Finally, we apply a principal component analysis (PCA) and k-means on the weighted matrix of the convolutional deep network to find clusters among architects.

### *4.1 Model accuracy*

For the DCNN classification task, top-$k$ error rates are important indicators in evaluating the performance of the model. Top 1 accuracy indicates the probability of whether the image can correctly match with the target label. Conversely, the top 5 accuracy suggests the probability of whether the correct image can appear with the target label among five pictures ordered according to their highest probability. The model was trained in 30 epochs and learning rate is set to 0.1 for the first 20 epochs and 0.01 for the last 10 epochs. Batch size is set to 16 images. The overall top 1 and top 5 training accuracy reached 99.7% and 100% respectively.

**Table 3.** Model accuracy for all categories

| Architect | Top1 accuracy | Top5 accuracy | Architect | Top1 accuracy | Top5 accuracy |
|---|---|---|---|---|---|
| A. Aalto | 65.07 | 82.53 | O. Niemeyer | 72.41 | 86.20 |
| A. Siza | 78.97 | 97.15 | P. Eisenman | 77.50 | 92.50 |
| B. Tschumi | 90.47 | 92.85 | R. Moneo | 82.85 | 91.42 |
| C. Himmelblau | 70.28 | 82.41 | R. Koolhaas | 32.60 | 63.04 |
| L. Corbusier | 67.12 | 82.19 | R. Piano | 66.23 | 89.61 |
| D. Libeskind | 76.36 | 85.45 | R. Meier | 79.03 | 90.32 |
| D. Perrault | 40.62 | 75.00 | R. Rogers | 62.50 | 92.85 |
| E. S. de Moura | 71.62 | 87.83 | SANNA | 77.08 | 87.50 |
| E. Miralles | 69.11 | 86.76 | S. Ban | 82.14 | 85.71 |
| F. Gehry | 80.80 | 95.95 | S. Holl | 48.48 | 69.69 |
| F. Lloyd Wright | 87.73 | 98.15 | T. Ando | 73.23 | 89.23 |
| F. Maki | 68.85 | 77.04 | K. Tange | 73.21 | 83.92 |
| I.M. Pei | 65.45 | 76.36 | T. Mayne | 77.77 | 80.00 |
| J. Nouvel | 71.15 | 90.38 | T. Ito | 56.79 | 92.59 |
| L. Kahn | 87.67 | 99.05 | Y. Taniguchi | 72.13 | 95.08 |
| M. van der Rohe | 84.42 | 95.90 | Z. Hadid | 65.51 | 83.90 |
| MVRDV | 60.00 | 88.57 | House | 79.78 | 93.14 |
| N. Foster | 77.41 | 90.32 | Total | 73.17 | 87.07 |

Table 3 shows the results of computing our model's top 1 accuracy and top 5 accuracy for architects. The average of the top 1 accuracy rate on the testing set is 73.2%, meaning that our model can predict the architect with this probability. The highest probabilities for top 1 accuracy are: Tschumi (90.4%), Lloyd Wright (87.7%), Kahn (87.6%), van der Rohe (84.4%), and Moneo (82.8%). Conversely, the lowest probabilities for top 1 accuracy are: Koolhaas (32.6%), Perrault (40.6%), Holl (48.4%), Ito (56.7%), and MVRDV (60.0%).

We can interpret these results as follows: The computer eye tends to be able to capture design features for the former group, which enables it to distinguish their architecture from others', but is likely to detect similar features for the latter group. Thus, the machine eye tends to confuse Koolhaas, Holl, Perrault with other architects, but it correctly distinguishes Kahn, Siza, and van der Rohe from others. And this tendency does not change if we focus on the top 5 accuracy: Kahn (99.0%), Lloyd Wright (98.1%), Siza (97.1%), van der Rohe (95.9%), and Gehry (95.9%) for the highest probabilities, and Koolhaas (63.0%), Holl (69.6%), Perrault (75.0%), Pei (76.3%), and Maki (77.0%) for the lowest ones. On average, almost 70% of architects can be distinguished with more than 80% probability if we focus on top 5 accuracy, and 45% of architects can be distinguished with more than 90% probability. In Kahn's case, this rises to 99.0%.

The result is intriguing, because we tend to consider that the characteristics of Koolhaas's and Holl's architecture lie in its unique material usage and form. On the other hand, Kahn's, Siza's, and van der Rohe's works are known as basic geometry-based designs, suggesting it would be easier to find more similarities between these architects. For example, van der Rohe is frequently considered to have established the design model for the office building, which is the rectangular appearance with multi-layers surrounded by the glass-curtain wall, making the landscape of our contemporary cities.

**Table 4.** Accuracy of different image types

| Image Source | Self-taken Images | | Internet Images | |
|---|---|---|---|---|
| Image Perspective | Indoor | Outdoor | Indoor | Outdoor |
| Top1 Accuracy | 70.72 | 66.19 | 74.72 | 73.84 |
| Top5 Accuracy | 85.24 | 81.25 | 89.90 | 88.13 |

We also examine whether or not there are significant differences that the computer vision captures between the indoor and outdoor images. Our result indicates that the indoor scenes are much more distinguishable to the machine eye than the outdoor ones (see Table 4). Although there may exist similar objects and features in the outdoor images (i.e., trees, pavements), the machine eye captures the characteristics of the interior spaces better than the external design features such as the form itself. Moreover, this indicates that the machine eye can find the characteristics of modern and contemporary architecture in the spatial design rather than the mass forms.

## 4.2 Grad-CAM

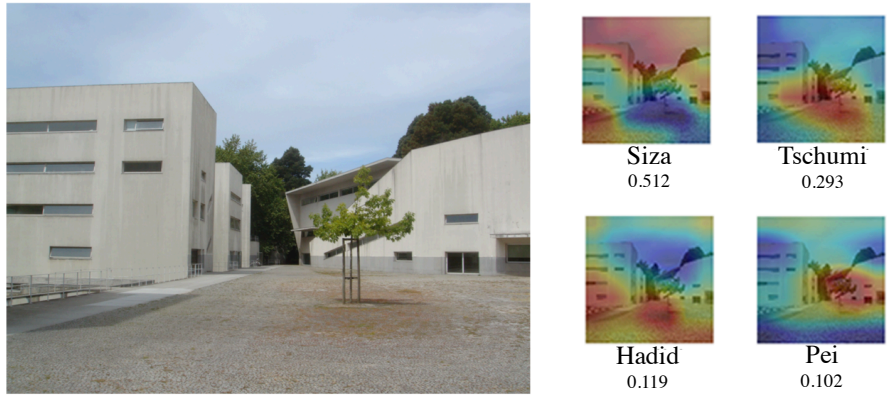

**Figure 4.** Image of Alvaro Siza's Porto school of architecture (left) and top 4 predictions (right).

Figure 4 shows an example of how the machine eye works by presenting Grad-CAM outputs. An exterior photo of Alvaro Siza's Porto school of architecture was fed into the trained model. The prediction of the top four categories is as follows: Siza, Tschumi, Hadid, and Pei. By using Grad-CAM, we were able to observe the evidence of the machine eye's focus on each image and the reason why the computer vision made the decision with the probability for each choice. In this example, we can observe that the building form was the main reason for the model to pick up Siza as its top choice. However, Pei's designs often have similar geometries; thus the model predicted Pei as its fourth choice and highlighted the similar area in the examined image.

### 4.3 Clustering by principal component analysis

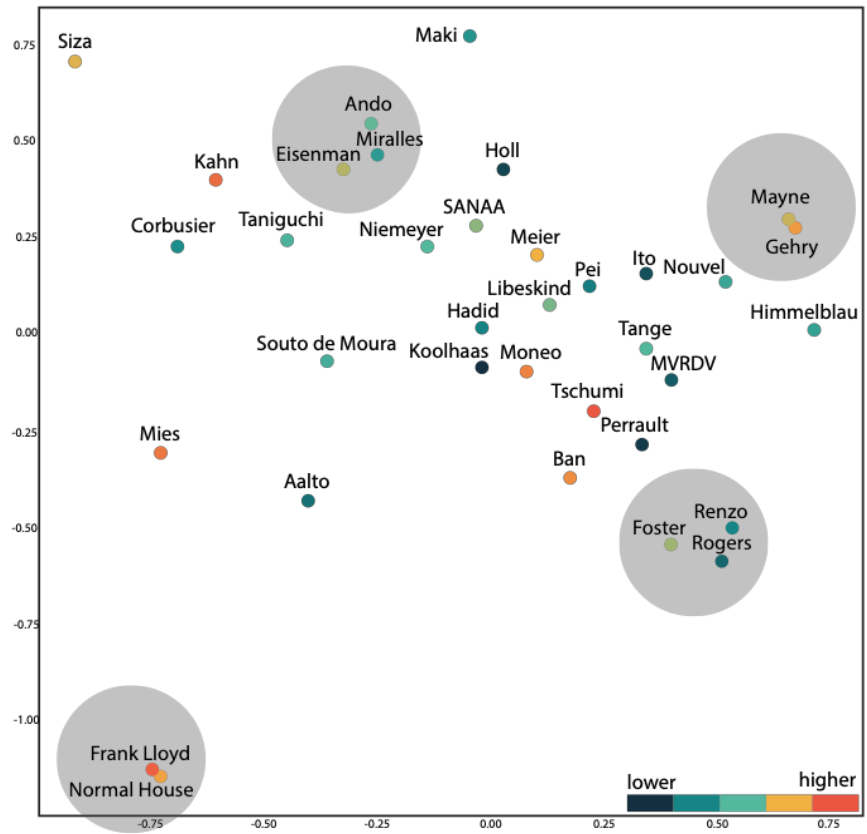

**Figure 5.** Linear PCA and clusters

Based on the results outlined in section 4.1, a clustering analysis was carried out using linear principal component analysis (see Figure 5). We measured the distance to judge similarities in architectural design between architects. Next, we reduced them using k-means to find clusters. Finally, we visualized the obtained results and observed four clusters of architects grouped by the machine. The following is our interpretation of the results.

The first cluster consists of Norman Forster, Richard Rogers, and Renzo Piano. They are frequently labeled in terms of "high-tech design" (Kron and Slesin, 1984), which pursues the expression of technology (i.e., structure and facilities) as design elements. They tend to borrow established technologies and materials from other fields (i.e., automobile or aircraft industry) and apply them to the construction process. Thus, their approach enables them to push the borders of architectural design. The development of the high-tech style is oriented to the eco-tech design or sustainable design architecture, which tries to reduce the environmental loads. The second cluster consists of Frank Lloyd Wright and "normal house." Most of his works are individual homes, although he designed more than 400 built works (800 if we include the unbuilt works). Wright established the "prairie style" at the beginning of his career, indicating the emphasis on horizontalness by lowering the height of the roof and using continuous windows and walls surrounding the building. This became the model for the middle-class suburban house in the U.S. and it rapidly spread over the entire country, resulting in the formation of urban and suburban landscapes.

The third cluster is made up of Frank Gehry and Thom Mayne (Morphosis). Both are based in Los Angeles, where digital technology and industrial materials provide their architectural characteristics. To generate the form, they start from the materials themselves and assemble those materials. Gehry's architecture is "to transform ordinary raw materials - unadorned chain link, sheet metal, glass, stucco and plywood- into essential formal elements of an intriguing architecture" (Stern, 1994, p.8), while Mayne overlaps several elements and expresses the incompleteness through his architecture. The fourth cluster consists of Enric Miralles, Peter Eisenman, and Tadao Ando. Although Miralles was not originally classified as a "deconstructivist" (Johnson and Wigley, 1988), the characteristics of his architecture can be described as fragmented, inclined roof and walls, seemingly inconstruction, which is similar to Eisenman's, who is classified as a "deconstructivist". Conversely, Ando's distinguishing features lie in severe geometric composition, together with exposed concrete and glass as materials, which seems to create a contrast with the other two architects. However, the Grad-CAM analysis gave us the insight that the machine eye captured curves circles as features of Ando's architecture, thus establishing a similarity with Miralles and Eisenman.

## 5. Discussion

This paper discusses the classification of architectural designs using the computer vision technique. We employed a deep convolutional neural network (DCNN) on a large-scale sample dataset of 34 architects and their architectural works. Our preliminary result provides an alternative view to the conventional classification methodology, i.e., architectural historians and theorists. Although it does not substitute conventional classifications, we show that our proposed methodology works rather as a complement to them and can shed light on unknown aspects of modern and contemporary architecture. The current study suggests the following contributions to the classification of an architect's design:

- Our algorithm enables us to identify an individual architect with 73% validity. We examined 34 architects from the different geographical areas and eras, including normal houses. This indicates that the trained neural network correctly captures the characteristics of an architect's design and differentiates them.

- The analysis of the model's acccuracy provides us with the differnece between the machine eye- and architectural historian and theorist-based classifications. While the computer eye can correctly classify Kahn, Siza, and van der Rohe with higher probabilities, it confuses Koolhaas and Holl with other architects. This indicates that the latter architects's design features cannot be detected by the computer, which is almost contrary to our intuition. Also, the prediction of the computer vision becomes more accurate for indoor scenes than outdoor ones, suggesting that the machine eye captures the visual features of the interior spatial design more precisely than the external form.

- Our analysis of the Grad-CAM of each architect identifies the design elements which differentiate architectural works. The visualization of this process enables us to uncover significant areas which the machine eye captures for the purpose of classification. For example, in the case of the Porto school of architecture by Alvaro Siza, the trained neural network identified the building form as Siza's design feature, resulting in a correct classification, but it also picked up Pei due to similar geometries.

- Most of our clustering analysis coincides with the conventional description made by architectural historians and theorists, indicating the validity of our methodology. The result shows that, for example, Forster, Rogers and Piano are successufully clustered as high-tech design. We also found that Wright is correctly clustered with the U.S. suburban house.

The proposed method provides clear value and novel perspectives to the existing research, but it also has several limitations. First, our sample size is small and varies greatly for each architect. Although the maximum number is more than 1,400, the minimum one is only around 200. These imbalanced categories may cause a potential bias for the analysis. Second, the current analysis does not consider the temporal factor, indicating that we do not distinguish an architect's work by his/her era. A specific architect's design is not necessarily consistent during his/her entire professional career; rather it changes due to new available technologies or social requirements. For example, Le Corbusier's former work is significantly different from his later one (i.e., Savoi, Ronchamp). A dataset considering the temporal factor would provide us with more insights on how some architects "grow together" or "grow apart" as time goes by. Finally, the number of the selected architects is also limited. We chose 34 architects as representatives, but the number should be increased for future work.

Any discussions on architecture requires the consideration of the social-economic-cultural movement behind the appearance of the design, its shape, and the employed technologies, because that design should be considered a consequence of that movement (Frampton, 1992). In addition, architecture is a spatial experience rather than a merely visual one. Thus, the most of the analysis and critique of modern and contemporary architecture emphasizes the necessity to consider human senses and factors such as memory, perception and cognition (Lynch, 1960; Rossi, 1960; Rowe, 1976). As a backdrop to these approaches, our approach measures visual similarities using a machine eye. This provides us with insights without considering any prior knowledge or any other human sensory information, which can be different from an analysis made by a human being. Thus, the current analysis can complement Kant's (1952) or Wolfflin's (1950), who analyze the aesthetics of spaces in terms of perception and discuss the cognitive process of architecture.

The application of a deep convolutional neural network (DCNN) in the context of architecture and urban planning would allow researchers to analyze visual similarities between types of architecture and create typologies and classifications of their design features. Although the methodology presented herein gives us preliminary results rather than complete ones, the method offers an effective means to analyze visual similarities and extract the features of an architect's design. It was difficult to separate and extract the visual factors from other factors such as prior knowledges, historical context, or personal imagination. This is a piece of critical information that was not obtainable prior to this study.